\documentclass[10pt,twocolumn,letterpaper]{article}

\usepackage{cvpr}      %

\usepackage{microtype}
\usepackage{inconsolata}

\usepackage{xcolor} %

\newcommand{\venue}[1]{{\footnotesize\color{gray}#1}}
\usepackage{xspace}
\newcommand{\ourmodel}{\mbox{WindowSeat}\xspace}

\newcommand{\bB}{\mathbf{B}}
\newcommand{\bT}{\mathbf{T}}
\newcommand{\bR}{\mathbf{R}}

\usepackage{makecell}
\usepackage{amssymb}%
\usepackage{pifont}%
\newcommand{\xmark}{\ding{55}}%

\usepackage{cuted}

\definecolor{cvprblue}{rgb}{0.21,0.49,0.74}
\usepackage[pagebackref,breaklinks,colorlinks,allcolors=cvprblue]{hyperref}

\def\paperID{970} %
\def\confName{CVPR}
\def\confYear{2026}

\title{
Reflection Removal through Efficient Adaptation of Diffusion Transformers 
}

\author{%
Daniyar Zakarin$^{*,1,2}$, 
Thiemo Wandel$^{*,2}$,
Anton Obukhov$^{\dagger,2}$,
Dengxin Dai$^{2}$\\
{\small%
$^{1}$ETH Zürich, 
$^{2}$HUAWEI Bayer Lab, 
$^{*}$Equal contributors, 
$^{\dagger}$Project lead
}
}

\begin{document}

\twocolumn[{%
\renewcommand\twocolumn[1][]{#1}%
\maketitle
\vspace{-1em}
\includegraphics[width=\linewidth]{figures/teaser_v07}
\captionof{figure}{
\textbf{We present WindowSeat, a model and fine-tuning protocol for one-step reflection removal.}
It repurposes a foundation image diffusion transformer (DiT) into a state-of-the-art computational photography tool, enabled by an efficient and scalable Physically Based Rendering (PBR) pipeline for data synthesis.
WindowSeat demonstrates stronger scene understanding and source-separation capabilities than competing methods, yielding cleaner outputs with fewer artifacts.
For the visualization above, the ``Scenario'', ``Ground truth'' transmission, and reflection layers were generated; 
``Photos with reflection'' were produced by our proposed PBR pipeline; 
results images are obtained from the respective methods. 
Best viewed zoomed in; 
arrows point at artifacts of methods;
contrast enhanced for visualization.
\vspace{3em}
}
\label{fig:teaser} 
}]

\begin{abstract}
We introduce a diffusion-transformer (DiT) framework for single-image reflection removal that leverages the generalization strengths of foundation diffusion models
in the restoration setting. 
Rather than relying on task-specific architectures, we repurpose a pre-trained DiT-based foundation model by conditioning it on reflection-contaminated inputs and guiding it toward clean transmission layers. 
We systematically analyze existing reflection removal data sources for diversity, scalability, and photorealism. 
To address the shortage of suitable data, we construct a physically based rendering (PBR) pipeline in Blender, built around the Principled BSDF, to synthesize realistic glass materials and reflection effects. 
Efficient LoRA-based adaptation of the foundation model, combined with the proposed synthetic data, achieves state-of-the-art performance on in-domain and zero-shot benchmarks. 
These results demonstrate that pretrained diffusion transformers, when paired with physically grounded data synthesis and efficient adaptation, offer a scalable and high-fidelity solution for reflection removal.
Project page: 
\href{https://hf.co/spaces/huawei-bayerlab/windowseat-reflection-removal-web}{https://hf.co/spaces/huawei-bayerlab/windowseat-reflection-removal-web}

\end{abstract}
    
\section{Introduction}
\label{sec:intro}
Reflections have long been persistent artifacts in photography, especially in casual or on-the-go capture. 
With most images now taken on mobile devices, and with glass and glossy materials increasingly common in modern architecture, it has become difficult to take a snapshot without including at least one unintended reflection.
Therefore, effective single-image reflection removal (SIRR) is essential for improving the quality of computational photography on mobile devices.
At its core, reflection removal is a fundamentally ill-posed source separation problem. 
Addressing it requires a strong scene understanding prior together with principles rooted in the physics of image formation.

Prior approaches often depended on task-specific architectures with limited pretraining~\cite{zhao2025reversible}, simplified physics-based assumptions such as modeling reflections as low-frequency components~\cite{li2014single}, and restricted data sources. 
These include screen-space simulation pipelines~\cite{Yang_2018_ECCV}, small-scale datasets~\cite{li2020single}, collections with potential pixel-level misalignment~\cite{zhu2024revisiting}, or pseudo-labels refined by manual post-processing~\cite{yang_openrr1k}.
This often resulted in sparse coverage of the underlying data distribution, leading to poor in-the-wild and out-of-distribution performance.

A natural way forward is to leverage a foundation model that already encodes rich scene understanding and fine-tune it for the task using only small amounts of in-domain data.
However, recent work on fine-tuning foundation models, both within their native domain~\cite{zhang2025scaling} and far outside it~\cite{marigoldaffordable,depthanythingv2}, shows that noisy or unrepresentative training data of any nature severely limits their ability to adapt while preserving the benefits of pretraining.
In contrast, sufficiently diverse photorealistic synthetic data provides the coverage and controllability needed to fully exploit these pretrained priors and adapt them to the target task.

This is the direction we take in this paper. 
Accordingly, we outline a practical path from foundation DiTs to strong single-image reflection removal solutions and contribute:
\begin{itemize}
\item a scalable PBR data pipeline that simulates light transport to produce realistic training data from large-scale sources;
\item an efficient protocol for adapting foundation DiTs, covering fine-tuning objectives, reuse of the base model's inputs and outputs, and practical settings for LoRA and quantization, enabling one-day training on a single consumer GPU and allowing future work to benefit from improved base models through a simple model swap;
\item a comprehensive evaluation of our model (\ourmodel) derived from a recent DiT on SIRR benchmarks and in-the-wild images, demonstrating top performance (Fig.~\ref{fig:teaser}).
\end{itemize}

\section{Related Work} 

\noindent \textbf{Taxonomy of reflection removal.}
To compensate for the missing prior knowledge about the scene, prior work often leverages side-channel information available only in controlled capture setups, such as 
flash-no-flash pairs~\cite{agrawal2005removing, wang2025flash}, 
polarization~\cite{nayar1997separation},
stereo baseline~\cite{niklaus2021learned},
or burst and video sequences~\cite{Liu_2020_CVPR}.
In contrast, our work advances the single-image (SIRR) setup with no special hardware support.

\noindent \textbf{Classical methods.}
Traditional reflection removal methods rely on statistical priors and simplified mathematical models of the reflection process \cite{NIPS2002_d5425997, guo2014robust, levin2004separating, gai2011blind}.
Modern end-to-end deep learning approaches surpass these classical ones, but depend on high-quality training data. 

\noindent \textbf{Real data.}
Existing real-world datasets are typically captured by placing a glass plate between the camera and the scene, collecting either single images \cite{SIR2-iccv17, zhang2018single, li2020single} or videos \cite{zhu2024revisiting, hu2025dereflection}.
However, these datasets are limited in the number of real scenes they can cover and often contain samples without pixel-perfect alignment due to refraction, object, or camera motion (e.g., wind-driven foliage or subtle tripod shifts).
Yang et al.~\cite{yang_openrr1k} propose a more scalable approach by starting from mobile captures with reflections and generating pseudo-ground truth using an existing reflection removal algorithm followed by manual refinement.
While easier to scale, this method is practical only for images requiring minimal manual cleanup and provides limited control over glass properties and the resulting data distribution.

\noindent \textbf{Alpha blending.}
Due to the limited availability of real-world reflection-removal datasets, many SIRR approaches additionally rely on synthetic data generated by alpha blending, typically created by overlaying images from PASCAL VOC~\cite{everingham2010pascal} \cite{hu2021trash, hu2023single, huang2025single, zhao2025reversible, hu2025dereflection, hong2024differ, wang2024promptrr} or COCO~\cite{lin2014microsoft} \cite{hu2025dereflection, hong2024differ, zhong2024language}.
These methods employ various screen-space mixing models, which lack the physical realism needed to capture true glass behavior.

\noindent \textbf{PBR data.}
Several methods have explored physical simulation as a source of high-quality training data. 
Kim et al.~\cite{kim2020single} reconstruct 3D meshes from RGBD images and simulate light transport between the scene and glass to reproduce anisotropic reflection effects. 
Meanwhile, \cite{slightrr} uses a carefully curated rendered dataset to learn the removal of both specular highlights and reflections.
Our approach extends this direction by providing a more scalable alternative in which the 3D scene remains fixed and only the input images vary, avoiding the need for per-scene geometric reconstruction while still preserving physically faithful light transport.

\noindent \textbf{Foundation models.}
Diffusion-based image generation~\cite{rombach2022high} has gained widespread attention across vision and graphics. 
Early text-to-image models often relied on auxiliary modules such as ControlNets~\cite{zhang2023adding} to enable practical fine-tuning.
For cross-modal prediction, Marigold~\cite{marigoldaffordable,ke2023repurposing} popularized latent concatenation with full-model fine-tuning.
LoRA~\cite{hu2022lora}, originally introduced as an efficient alternative to full fine-tuning in transformer-heavy NLP, became the default strategy for adapting diffusion models once they evolved into DiTs, where full fine-tuning is prohibitively expensive.
As we show in this paper, modern flow matching~\cite{lipman2022flow} image-editing models~\cite{batifol2025flux} are sufficiently expressive to maintain high image quality with minimal modification, even under lightweight 4-bit QLoRA adaptation~\cite{liu2025fluxqlora}.

\begin{figure*}[t!]
\centering
\includegraphics[width=\textwidth]{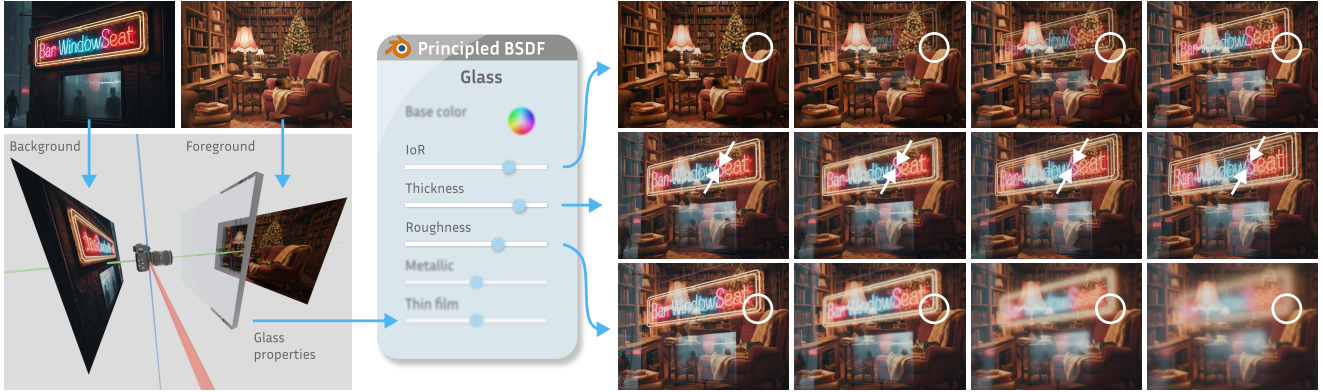}
\caption{
\textbf{Physically Based Rendering (PBR) pipeline for synthetic data generation.}
\textit{Left:}
The synthesis begins by sampling the foreground and background images, which can be in sRGB or HDR formats.
The images are placed into a static 3D scene with a glass plate positioned in front of a virtual camera.
The camera parameters and object distances are chosen to cover the view frustum of the virtual camera along transmission and reflection paths.
\textit{Middle:}
At the heart of our pipeline is the Principled BSDF shading model~\cite{Burley2012PhysicallyBasedSA, 2015ExtendingTD}, 
implemented in Blender~\cite{blender}, which enables simulation of a wide range of photorealistic glass effects and light interactions. 
\textit{Right:}
Visualizations of three factors of variation. 
Index of Refraction (IoR) affects reflection strength.
Thickness increases ghosting, which appears as larger gap between the multiple reflections (arrows).
Roughness controls the degree of scatter and blur.
Such a simulation cannot be faithfully reproduced by screen-space alpha blending models.
Details in Sec.~\ref{sec:alpha_blending} and \ref{sec:data_synthesis}.
Best viewed zoomed in.
}
\label{fig:pbr}
\end{figure*}

\noindent \textbf{Foundation models and SIRR.}
To exploit potential synergies between the transmission and reflection layers, several works predict both modalities simultaneously~\cite{hu2021trash, hu2023single, hu2024single, huang2025single, zhao2025reversible, zhong2024language}.
Recent diffusion-based approaches further extend SIRR with multi-step~\cite{hong2024differ, zhong2024language, wang2024promptrr} or single-step~\cite{hu2025dereflection} denoising, and some incorporate text encoders to provide semantic priors~\cite{hong2024differ, zhong2024language, wang2024promptrr}. 
However, these methods rely on costly multi-step sampling, auxiliary ControlNets, cross-latent skip connections, or cross-attention for dual-stream prediction, all of which fall outside our goal of a simple and efficient fine-tuning protocol.
Instead, we adapt an image-editing diffusion model~\cite{batifol2025flux} into a feed-forward reflection-removal network.
The procedure for obtaining \ourmodel remains straightforward, and the resulting model outperforms prior work across the board.

\label{sec:relwork}

\section{Method}
\label{sec:method}

\subsection{Recap: Alpha Blending}
\label{sec:alpha_blending}
A central obstacle in developing robust reflection-removal systems is the acquisition of training data. Capturing real-world transmission and reflection-contaminated image pairs is difficult due to refractive pixel shifts, inconsistent illumination, and scene or camera motion. Consequently, most existing methods 
predominantly rely on synthetic data, typically generated using some form of a linear blending of two images to simulate glass reflections over natural scenes. Recent state of the art methods~\cite{hu2025dereflection, hu2024single, cai2025f2t2, zhao2025reversible} adopt the formulation introduced in \cite{hu2023single}:
\[
    \bB = \alpha \bT + \beta \bR - \alpha \beta \bT \circ \bR,
\]
where $\bB$ denotes the blended output, and $\bT$ and $\bR$ represent the transmission and reflection images, respectively. We refer to this approach as \emph{alpha blending}. 

Although simple and computationally efficient, this approach fails to capture the underlying optical phenomena responsible for realistic reflections, such as subsurface scattering, which produces characteristic blurring, and multiple internal reflections, which give rise to ghosting artifacts in real glass. In practice, Gaussian blurring is often used to approximate the scattering, which provides limited realism. Moreover, alpha blending operates in the standard RGB (sRGB) color space and cannot produce high-intensity specular highlights on glass surfaces. 

\subsection{Physically Based Rendering Pipeline}
\label{sec:data_synthesis}
To address these limitations of alpha blending, we propose a light-weight \textbf{physically based rendering (PBR)} pipeline that explicitly simulates the light-glass interaction (Fig.~\ref{fig:pbr}). We achieve this by capturing a transmission scene with simulated reflections from a glass surface, where the reflection source is either a panoramic high-dynamic-range (HDR) environment map or a planar RGB image. HDR panoramic images are widely used in 3D modeling, and in our case they help us generate light interaction in a wider light spectrum, generating strong effects like hazy scattering or strong highlights. The panoramic format enables modeling of light from multiple incident directions. In contrast, the wide availability of RGB images as planar reflection sources, such as COCO~\cite{lin2014microsoft} and Pascal VOC~\cite{everingham2010pascal}, lets us generate a rich family of reflection patterns. 

For each training sample, the virtual camera is positioned to observe a scene through a glass material. For efficiency, each scene is represented as a textured mesh plane, which greatly accelerates rendering while preserving spatial detail. Unlike sRGB images, HDR sources encode radiance rather than display intensities, enabling more physically accurate reflection synthesis and internal light interactions.

To model the glass material, we use Blender’s Principled BSDF shader~\cite{blender} (Fig.~\ref{fig:pbr}), a physically based shading model derived from Disney’s BRDF~\cite{Burley2012PhysicallyBasedSA, 2015ExtendingTD}. This formulation provides a compact parameterization of the optical properties relevant to glass. The index of refraction (IoR) and metallic parameters modulate the reflection strength, while the surface roughness controls the microfacet distribution that determines the blur of specular reflections. Light attenuation through the medium is simulated by assigning a base color to the material, and ghosting is produced by varying the glass thickness. Reflection-free ground-truth images are rendered by setting IoR to 1.0 and zeroing the metallic and roughness terms. By carefully controlling the glass thickness and the camera–glass spacing, we avoid refractive pixel shifts and maintain pixel alignment between the reflection-contaminated and transmission images.

These controls enable systematic ablation studies (Sec.~\ref{sec:ablation}) and flexible synthesis of training data.

\begin{figure}[t!]
\centering
\resizebox{\linewidth}{!}{
\includegraphics[width=\textwidth]{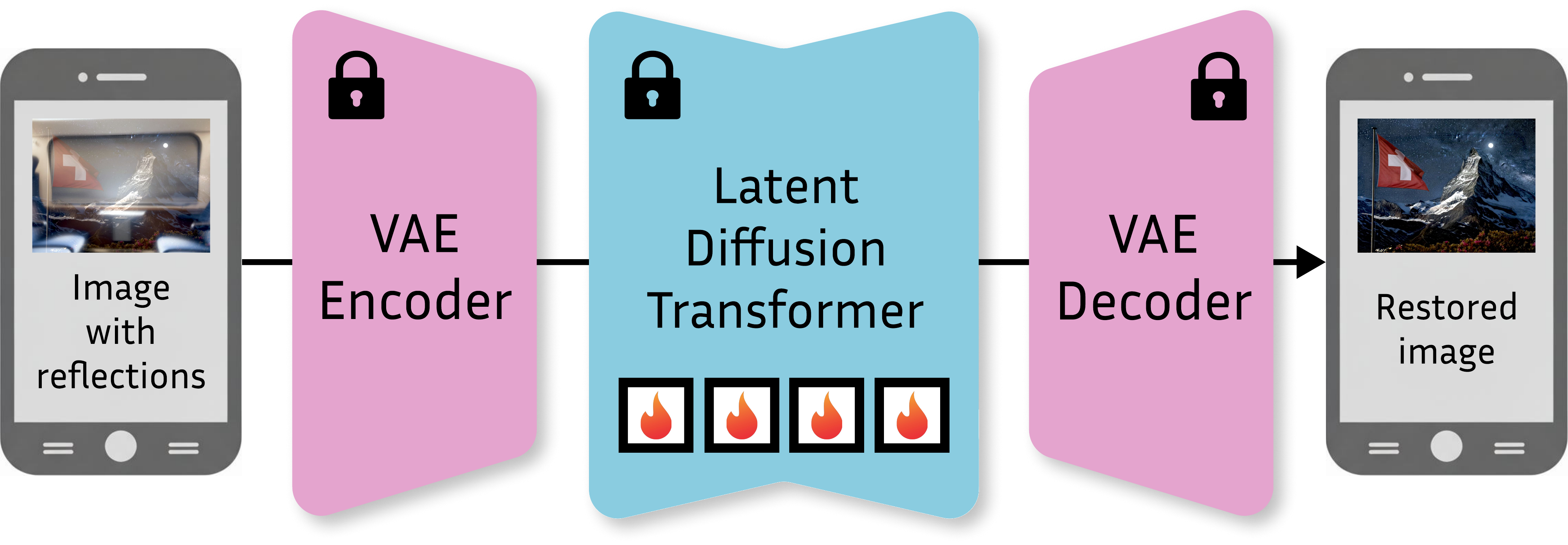}
}
\caption{
\textbf{Model architecture.} 
Foundation DiTs~\cite{peebles2023scalable} operate in a compressed latent space in the bottleneck of a VAE~\cite{Kingma2014}.
Fine-tuning DiTs can be done efficiently with lightweight LoRA~\cite{hu2022lora} adapters.
Modern DiTs~\cite{batifol2025flux} with more than 10B parameters often employ quantized representations, such as QLoRA~\cite{qlora, liu2025fluxqlora}.
The end-to-end fine-tuning procedure is elaborated in Sec.~\ref{sec:one_step_fm_rr}.
}
\label{fig:arch}
\end{figure}

\subsection{One-Step Flow Matching}
\label{sec:one_step_fm_rr}

We perform reflection removal in the latent space of a frozen variational autoencoder (VAE). 
Given a reflection-contaminated input image $\mathbf{B}$ and clean target $\mathbf{T}$, we encode
$\mathbf{z}_{\!B}=\mathcal{E}(\mathbf{B})$ with encoder $\mathcal{E}$ and decode with the frozen decoder $\mathcal{D}$ (see Fig.~\ref{fig:arch}).
Our DiT backbone was pre-trained with a flow matching objective and processes two latent token streams along with a text prompt.
Concretely, it operates on two sequences of latent tokens of identical spatial layout: a primary stream containing the encoded input latent $\mathbf{z}_{B}$ and an auxiliary second stream.
In pre-training, the second stream receives a stochastically perturbed ground truth image latent, whereas in our setup we \emph{duplicate} the latent, \ie $\tilde{\mathbf{z}}_{B} = \mathbf{z}_{B}$ (see ablation in Sec.~\ref{sec:ablation}).
The text embedding is precomputed in advance.

Let $v_\theta(\cdot\,;\mathbf{p})$ be the DiT-predicted latent \emph{velocity} under prompt conditioning $\mathbf{p}$. 
Our one-step update produces a reflection-free latent
{%
\[
    \hat{\mathbf{z}}_{\text{edit}}
    \;=\;
    \mathbf{z}_{\!B} \;+\; v_\theta(\mathbf{z}_{\!B}\,;\mathbf{p}),
    \qquad
    \hat{\mathbf{Y}} \;=\; \mathcal{D}\!\left(\hat{\mathbf{z}}_{\text{edit}}\right),
\]
}

\noindent replacing multi-step diffusion sampling with a \emph{single} forward pass while leveraging the DiT pre-training.
Unlike prior diffusion pipelines, we do not use ControlNet modules, cross-latent skip connections, losses in latent space, or any VAE fine-tuning. We train the DiT LoRA adapters in pixel space with PSNR and SSIM losses between $\hat{\mathbf{Y}}$ and $\mathbf{T}$:
{%
\[
\mathcal{L}
= \lambda_{\text{PSNR}}\,\mathcal{L}_{\text{PSNR}}(\hat{\mathbf{Y}}, \mathbf{T})
+ \lambda_{\text{SSIM}}\,\mathcal{L}_{\text{SSIM}}(\hat{\mathbf{Y}}, \mathbf{T}).
\]
}

\section{Experiments}
\label{sec:experiments}

\begin{table*}[t] %
\centering
\caption{\textbf{Quantitative comparison} on in-domain (\textit{Nature} ~\cite{li2020single}, \textit{Real} ~\cite{zhang2018single}) and zero-shot (\textit{SIR}$^2$ ~\cite{SIR2-iccv17}) datasets. Values are sourced from the respective papers, unpublished values are denoted with `--'. \textit{SIR}$^2$ (454) is the weighted average of \textit{Objects} (200), \textit{Postcard} (199), and \textit{Wild} (55). Best viewed on screen and zoomed in. }
\label{tab:main_benchmark}
\setlength{\tabcolsep}{4pt}
\renewcommand{\arraystretch}{1.12}
\begin{adjustbox}{max width=\textwidth} %
\begin{tabular}{l *{14}{c}}
\toprule
\multirow{2}{*}{\raisebox{-0.5ex}{Method}} &
\multicolumn{2}{c}{\textit{Nature} ~\cite{li2020single} (20)} &
\multicolumn{2}{c}{\textit{Real} ~\cite{zhang2018single} (20)} &
\multicolumn{2}{c}{\textit{Objects} ~\cite{SIR2-iccv17} (200)} &
\multicolumn{2}{c}{\textit{Postcard} ~\cite{SIR2-iccv17} (199)} &
\multicolumn{2}{c}{\textit{Wild} ~\cite{SIR2-iccv17} (55)} &
\multicolumn{2}{c}{\textit{SIR}$^2$ ~\cite{SIR2-iccv17} (454)} &
\multicolumn{2}{c}{\textit{SIR}$^2$ ~\cite{SIR2-iccv17} (500)} \\
\cmidrule(lr){2-3} \cmidrule(lr){4-5} \cmidrule(lr){6-7} \cmidrule(lr){8-9} \cmidrule(lr){10-11} \cmidrule(lr){12-13} \cmidrule(lr){14-15}
 & PSNR$\uparrow$ & SSIM$\uparrow$
 & PSNR$\uparrow$ & SSIM$\uparrow$
 & PSNR$\uparrow$ & SSIM$\uparrow$
 & PSNR$\uparrow$ & SSIM$\uparrow$
 & PSNR$\uparrow$ & SSIM$\uparrow$
 & PSNR$\uparrow$ & SSIM$\uparrow$ 
 & PSNR$\uparrow$ & SSIM$\uparrow$ \\
\midrule
DSRNet (extra data) \cite{hu2023single} \venue{(ICCV 2023)} 
& -- & -- & 23.91 & 0.818 & 26.74 & 0.920 & 24.83 & 0.911 & 26.11 & 0.906 & 25.83 & 0.914 & -- & -- \\
DSRNet (w/o extra data) \cite{hu2023single} \venue{(ICCV 2023)}
& -- & -- & 24.23 & 0.820 & 26.28 & 0.914 & 24.56 & 0.908 & 25.68 & 0.896 & 25.45 & 0.909 & -- & -- \\
RRW \cite{zhu2024revisiting} \venue{(CVPR 2024)}
& 25.96 & 0.843 & 23.82 & 0.817 & -- & -- & -- & -- & -- & -- & 25.45 & 0.910 & -- & -- \\ 
L-DiffER \cite{hong2024differ} \venue{(ECCV 2024)}
& 23.95 & 0.831 & 23.77 & 0.821 & -- & -- & -- & -- & -- & -- & -- & -- & 25.18 & 0.911 \\
DSIT (data setting I) \cite{hu2024single} \venue{(NeurIPS 2024)}
& -- & -- & 25.06 & 0.836 & 26.81 & 0.919 & 25.63 & 0.924 & 27.06 & 0.910 & 26.32 & 0.920 & -- & -- \\ 
DSIT (data setting II) \cite{hu2024single} \venue{(NeurIPS 2024)}
& 26.77 & 0.847 & 25.22 & 0.836 & 27.27 & 0.932 & 25.58 & 0.922 & 27.40 & 0.918 & 26.54 & 0.926 & -- & -- \\ 
RDNet w/o nature \cite{zhao2025reversible} \venue{(CVPR 2025)}
& -- & -- & 24.43 & 0.835 & 25.76 & 0.905 & 25.95 & 0.920 & 27.20 & 0.910 & 26.02 & 0.912 & -- & -- \\ 
RDNet w nature \cite{zhao2025reversible} \venue{(CVPR 2025)}
& -- & -- & 25.58 & 0.846 & 26.78 & 0.921 & 26.33 & 0.922 & 27.70 & 0.915 & 26.69 & 0.921 & -- & -- \\
F2T2-HiT \cite{cai2025f2t2} \venue{(ICIP 2025)}
& 26.08 & 0.837 & 21.64 & 0.766 & -- & -- & -- & -- & -- & -- & 25.72 & 0.903 & -- & -- \\ 
Huang et al. \cite{huang2025single} \venue{(arXiv 2025)}
& 27.03 & \underline{0.853} & 25.12 & 0.828 & 27.07 & 0.930 & 26.43 & 0.931 & 27.96 & 0.922 & 26.90 & 0.929 & -- & -- \\ 
DAI \cite{hu2025dereflection} \venue{(arXiv 2025)}
& 26.81 & 0.843 & 25.21 & 0.841 & -- & -- & -- & -- & -- & -- & -- & -- & 27.19 & 0.930 \\ 
\midrule 
\ourmodel (ours) %
& \underline{27.12} & 0.849 & \underline{26.28} & \underline{0.856} & \underline{28.81} & \textbf{0.944} & \textbf{29.17} & \textbf{0.934} & \underline{28.97} & \underline{0.935} & \textbf{28.99} & \textbf{0.939} & \textbf{28.75} & \textbf{0.940} \\
 \ourmodel (ours, Apache 2.0) %
& \textbf{27.57} & \textbf{0.855}
& \textbf{26.60} & \textbf{0.864}
& \textbf{28.85} & \underline{0.938}
& \underline{28.70} & \underline{0.933}
& \textbf{29.44} & \textbf{0.936}
& \underline{28.84} & \underline{0.936}
& \underline{28.60} & \underline{0.937} \\

\bottomrule
\end{tabular}
\end{adjustbox}
\end{table*}

\begin{table*}[t]
\centering
\caption{\textbf{Quantitative comparison} with MS-SSIM$\uparrow$ and LPIPS$\downarrow$ evaluation metrics. Values are re-evaluated by saving all predictions as images on disk and computing metrics in 8-bit precision. \textit{Nature} (20) and \textit{Real} (20) are in-domain, while \textit{Objects} (200), \textit{Postcard} (199), and \textit{Wild} (55) are evaluated in a zero-shot manner.}
\label{tab:offline_evaluation}
\setlength{\tabcolsep}{4pt}
\renewcommand{\arraystretch}{1.12}
\begin{adjustbox}{max width=\textwidth}
\begin{tabular}{l cc cc cc cc cc}
\toprule
\multirow{2}{*}{\raisebox{-0.5ex}{Method}} &
\multicolumn{2}{c}{\textit{Nature} (20)} &
\multicolumn{2}{c}{\textit{Real} (20)} &
\multicolumn{2}{c}{\textit{Objects} (200)} &
\multicolumn{2}{c}{\textit{Postcard} (199)} &
\multicolumn{2}{c}{\textit{Wild} (55)} \\
\cmidrule(lr){2-3} \cmidrule(lr){4-5} \cmidrule(lr){6-7} \cmidrule(lr){8-9} \cmidrule(lr){10-11}
 & MS-SSIM$\uparrow$ & LPIPS$\downarrow$
 & MS-SSIM$\uparrow$ & LPIPS$\downarrow$
 & MS-SSIM$\uparrow$ & LPIPS$\downarrow$
 & MS-SSIM$\uparrow$ & LPIPS$\downarrow$
 & MS-SSIM$\uparrow$ & LPIPS$\downarrow$ \\
\midrule
DSRNet \cite{hu2023single} \venue{(ICCV 2023)}
& 0.9144 & 0.1478
& 0.8737 & 0.1831
& 0.9564 & 0.0847
& 0.9263 & 0.1260
& 0.9338 & 0.1096 \\
DAI \cite{hu2025dereflection} \venue{(arXiv 2025)}
& 0.9309 & 0.2161
& 0.9045 & 0.1790
& 0.9638 & 0.0689
& 0.9567 & 0.1029
& 0.9423 & 0.0941 \\
RDNet \cite{zhao2025reversible} \venue{(CVPR 2025)}
& 0.9231 & \textbf{0.1361}
& 0.9081 & 0.1442
& 0.9609 & 0.0836
& 0.9361 & 0.1121
& 0.9406 & 0.0992 \\
DSIT \cite{hu2024single} \venue{(NeurIPS 2024)}
& 0.9223 & 0.1598
& 0.8934 & 0.1618
& 0.9586 & 0.0939
& 0.9441 & 0.1242
& 0.9447 & 0.0967 \\
\midrule
\ourmodel (ours)
& \underline{0.9435} & \underline{0.1368}
& \underline{0.9296} & \underline{0.1131}
& \underline{0.9759} & \textbf{0.0470}
& \textbf{0.9693} & \textbf{0.0504}
& \underline{0.9625} & \textbf{0.0632}
 \\
 \ourmodel (ours, Apache 2.0) 
& \textbf{0.9494} & \textbf{0.1355}
& \textbf{0.9396} & \textbf{0.1074}
& \textbf{0.9661} & \underline{0.0550}
& \underline{0.9664} & \underline{0.0549}
& \textbf{0.9655} & \underline{0.0682} \\
\bottomrule
\end{tabular}
\end{adjustbox}
\end{table*}

\begin{table}[ht]
\caption{
\textbf{Parameter count} per component. 
We train 3.6\% of all parameters and quantize 95.7\% of them to 4-bit. 
The text encoder is not used at inference. 
The total model size is $\sim$12.5B parameters.
}
\label{tab:param_counts_compact}
\setlength{\tabcolsep}{4pt}
\renewcommand{\arraystretch}{1.1}
\begin{adjustbox}{max width=\linewidth}
\begin{tabular}{lrrrr}
\toprule
Component & Total Params (K) & 16-bit (K) & 4-bit (K) & Trainable (K) \\
\midrule 
VAE Encoder                      & 34,274      & 34,274      & 0               & 0 \\
DiT (w/o LoRA) & 11,901,408  & 396         & 11,901,012  & 0 \\
LoRA Adapter                     & 450,249     & 450,249     & 0               & 450,249 \\
VAE Decoder                      & 49,545      & 49,545      & 0               & 0 \\
\midrule
\textbf{Total}                   & \textbf{12,435,477} & \textbf{534,465} & \textbf{11,901,012} & \textbf{450,249} \\
\bottomrule
\end{tabular}
\end{adjustbox}
\end{table}

\begin{table*}[t]
\centering
\caption{\textbf{Ablation on DiT input choices.} Mode indicates the output parameterization: FLOW predicts a latent-space
velocity $\mathbf{v}_\theta$ that is added to the image latent
$\mathbf{z}_{\!B}=\mathcal{E}(\mathbf{B})$, whereas LATENT predicts the edited
latent $\hat{\mathbf{z}}_{\text{edit}}$ directly.
$z_1$ and $z_2$ denote the first and second latent token sets fed to the DiT;
$\mathcal{N}(0,\mathbf{I})$
is Gaussian noise and ``---'' indicates that no second latent is used.
}
\label{tab:ablation_pred_z1_z2}
\setlength{\tabcolsep}{4pt}
\renewcommand{\arraystretch}{1.12}
\begin{adjustbox}{max width=0.95\textwidth}
\begin{tabular}{c c c *{10}{c}}
\toprule
\multirow{2}{*}{\raisebox{-0.5ex}{Mode}} & \multirow{2}{*}{\raisebox{-0.5ex}{$z_1$}} & \multirow{2}{*}{\raisebox{-0.5ex}{$z_2$}} &
\multicolumn{2}{c}{\textit{Nature} (20)} &
\multicolumn{2}{c}{\textit{Real} (20)} &
\multicolumn{2}{c}{\textit{Postcard} (199)} &
\multicolumn{2}{c}{\textit{Objects} (200)} &
\multicolumn{2}{c}{\textit{Wild} (55)} \\
\cmidrule(lr){4-5}\cmidrule(lr){6-7}\cmidrule(lr){8-9}\cmidrule(lr){10-11}\cmidrule(lr){12-13}
\multicolumn{3}{c}{} &
PSNR $\uparrow$ & SSIM $\uparrow$ &
PSNR $\uparrow$ & SSIM $\uparrow$ &
PSNR $\uparrow$ & SSIM $\uparrow$ &
PSNR $\uparrow$ & SSIM $\uparrow$ &
PSNR $\uparrow$ & SSIM $\uparrow$ \\
\midrule
\textsc{Flow}   & $\mathcal{E}(\mathbf{B})$ & \textemdash &
26.18 & 0.816 &
24.68 & 0.801 &
26.06 & 0.863 &
26.47 & 0.890 &
27.10 & 0.889 \\
\textsc{Flow}   & $\mathcal{E}(\mathbf{B})$ & $\mathcal{N}(0, \mathbf{I})$  &
\textbf{27.43} & \textbf{0.849} &
\underline{26.19} & \textbf{0.861} &
\underline{28.58} & \underline{0.929} &
\underline{28.46} & \textbf{0.937} &
\textbf{28.98} & \underline{0.931} \\
\textsc{Latent} & $\mathcal{E}(\mathbf{B})$ & $\mathcal{E}(\mathbf{B})$ &
26.58 & \underline{0.837} &
25.43 & 0.824 &
28.18 & 0.924 &
27.89 & 0.926 &
27.48 & 0.908 \\
\textsc{Flow} & $\mathcal{E}(\mathbf{B})$ & $\mathcal{E}(\mathbf{B})$ &
\underline{27.12} & \textbf{0.849} &
\textbf{26.28} & \underline{0.856} &
\textbf{28.81} & \textbf{0.944} &
\textbf{29.17} & \underline{0.934} &
\underline{28.97} & \textbf{0.935} \\
\bottomrule
\end{tabular}
\end{adjustbox}
\end{table*}

\begin{table*}[t]
\centering
\caption{\textbf{Ablation on loss terms} evaluated on \textit{Nature} (20), \textit{Real} (20), and \textit{SIR}$^2$ sub-datasets (\textit{Postcard} (199), \textit{Objects} (200), \textit{Wild} (55)).} 
\label{tab:ablation_losses}
\setlength{\tabcolsep}{4pt}
\renewcommand{\arraystretch}{1.12}
\begin{adjustbox}{max width=0.85\textwidth}
\begin{tabular}{cc *{10}{c}}
\toprule
\multirow{2}{*}{\raisebox{-0.5ex}{$\mathcal{L}_{\mathrm{SSIM}}$}} &
\multirow{2}{*}{\raisebox{-0.5ex}{$\mathcal{L}_{\mathrm{PSNR}}$}} &
\multicolumn{2}{c}{\textit{Nature} (20)} &
\multicolumn{2}{c}{\textit{Real} (20)} &
\multicolumn{2}{c}{\textit{Objects} (200)} &
\multicolumn{2}{c}{\textit{Postcard} (199)} &
\multicolumn{2}{c}{\textit{Wild} (55)} 
\\
\cmidrule(lr){3-4}\cmidrule(lr){5-6}\cmidrule(lr){7-8}\cmidrule(lr){9-10}\cmidrule(lr){11-12}
\multicolumn{1}{c}{} & \multicolumn{1}{c}{} &
PSNR $\uparrow$ & SSIM $\uparrow$
& PSNR $\uparrow$ & SSIM $\uparrow$
& PSNR $\uparrow$ & SSIM $\uparrow$
& PSNR $\uparrow$ & SSIM $\uparrow$
& PSNR $\uparrow$ & SSIM $\uparrow$ \\
\midrule
\xmark & \checkmark
& \textbf{27.65} & \underline{0.847}
& \textbf{26.40} & \underline{0.850}
& \underline{28.75} & 0.928
& \underline{28.87} & \textbf{0.940}
& \underline{28.89} & \underline{0.934} \\
\checkmark & \xmark
& 23.16 & 0.805
& 22.90 & 0.785
& 28.31 & \underline{0.930}
& 23.08 & 0.872
& 21.62 & 0.830 \\
\midrule
\checkmark & \checkmark
& \underline{27.12} & \textbf{0.849}
& \underline{26.28} & \textbf{0.856}
& \textbf{28.81} & \textbf{0.944}
& \textbf{29.17} & \underline{0.934}
& \textbf{28.97} & \textbf{0.935} \\
\bottomrule
\end{tabular}

\end{adjustbox}
\end{table*}

\begin{figure}[t!]
\centering
\includegraphics[width=\linewidth]{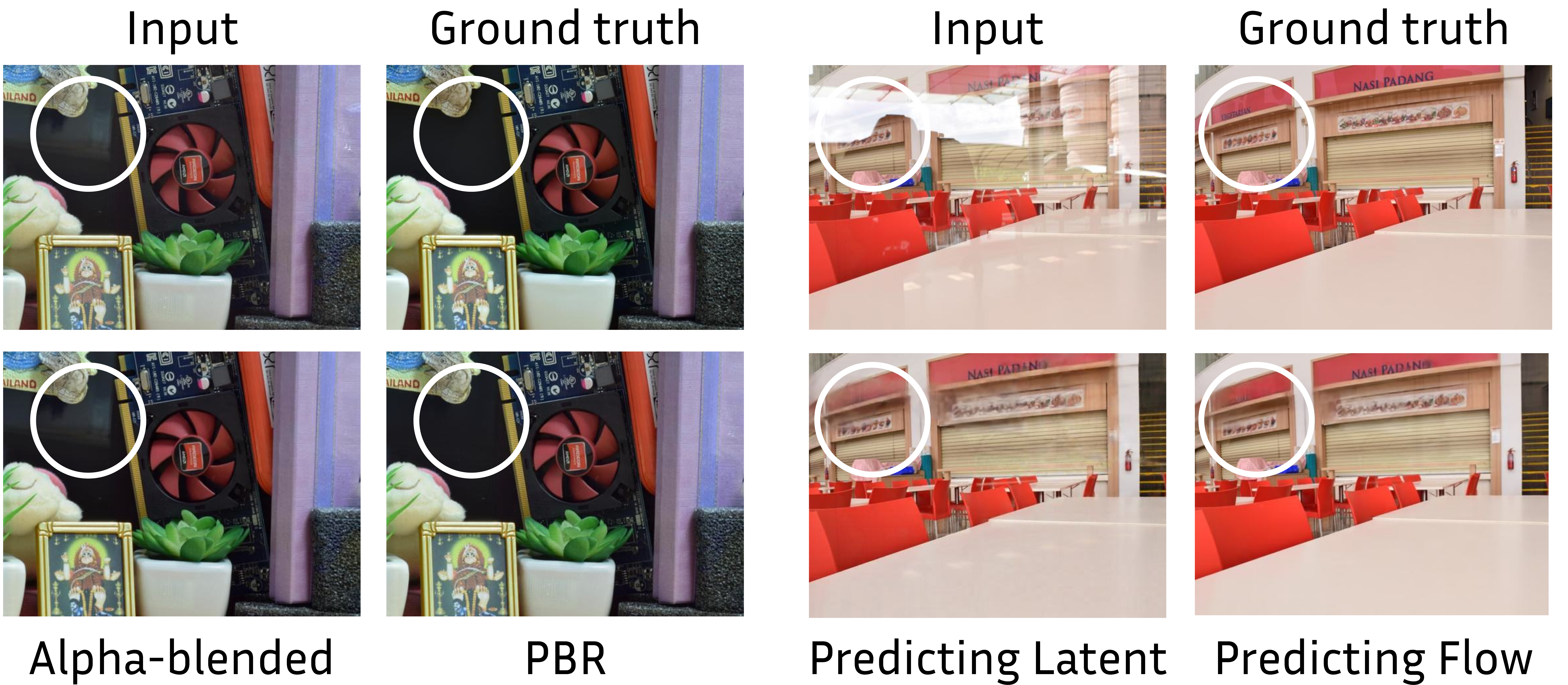}
\caption{
\textbf{Ablations}.
\textit{Left:} 
PBR vs Alpha Blending;
\textit{Right:}
Latent vs Flow objectives (Sec.~\ref{sec:ablation}).
As seen, PBR data and Flow objective produce more accurate results.
Best viewed zoomed in. 
}
\label{fig:ablation_pbr_and_obj}
\end{figure}

\newcommand{\rowH}{23.5mm}   %
\newcommand{\colgap}{3pt}  %
\newcommand{\textgap}{5pt} %
\newcommand{\labelsize}{\large} %

\begin{figure*}[ht!] %
\centering

\setlength{\tabcolsep}{0pt}
\begin{tabular}
{@{}p{0pt}@{\hspace{\textgap}}c@{\hspace{\colgap}}c@{\hspace{\colgap}}c@{\hspace{\colgap}}c@{\hspace{\colgap}}c@{}}

\parbox[c][\rowH][c]{0pt}{\makebox[0pt][r]{\labelsize\rotatebox{90}{Input}}} &
  \raisebox{-0.5\height}{\includegraphics[height=\rowH]{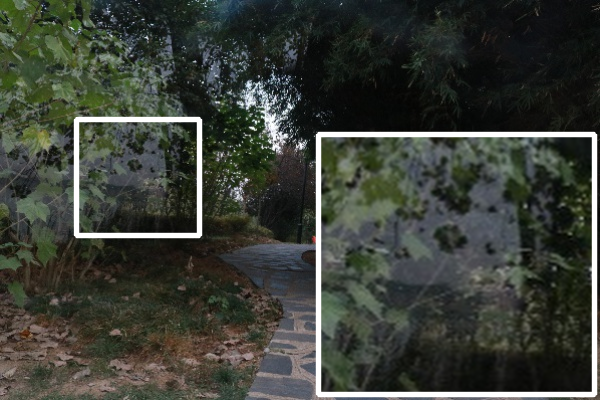}} &
  \raisebox{-0.5\height}{\includegraphics[height=\rowH]{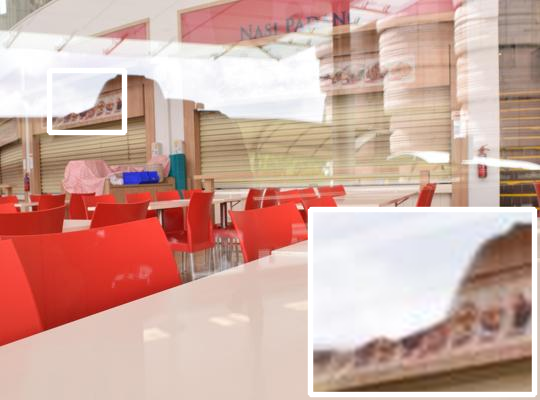}} &
  \raisebox{-0.5\height}{\includegraphics[height=\rowH]{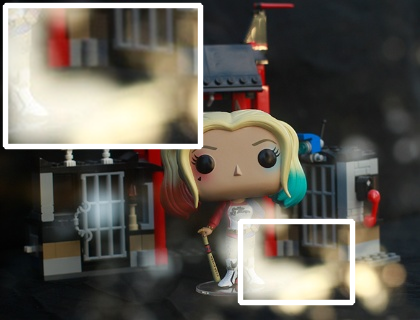}} &
  \raisebox{-0.5\height}{\includegraphics[height=\rowH]{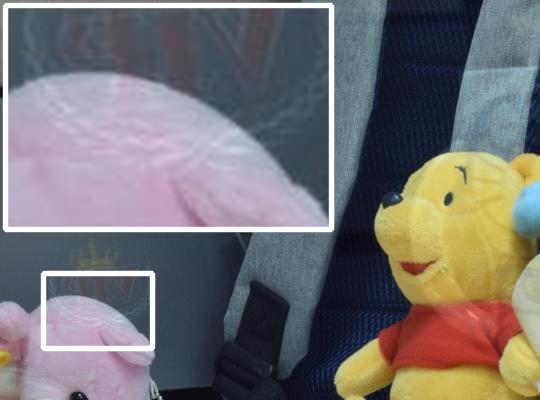}} &
  \raisebox{-0.5\height}{\includegraphics[height=\rowH]{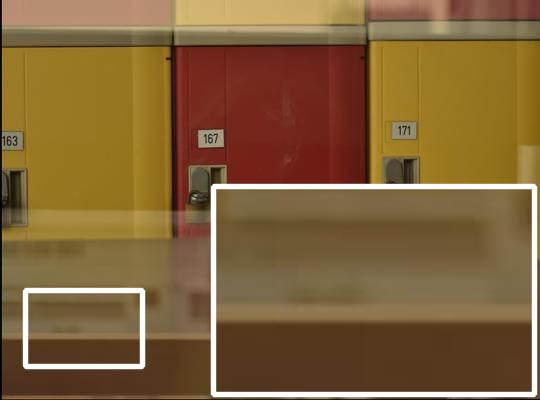}} \\
  
\parbox[c][\rowH][c]{0pt}{\makebox[0pt][r]{\labelsize\rotatebox{90}{GT}}} &
  \raisebox{-0.5\height}{\includegraphics[height=\rowH]{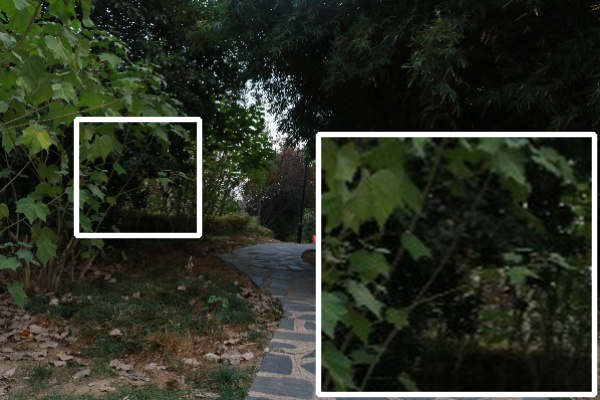}} &
  \raisebox{-0.5\height}{\includegraphics[height=\rowH]{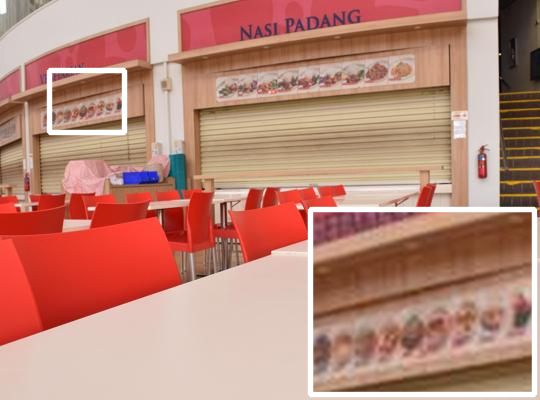}} &
  \raisebox{-0.5\height}{\includegraphics[height=\rowH]{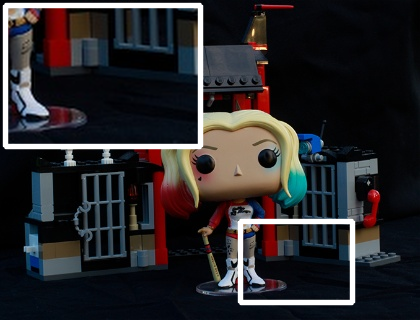}} &
  \raisebox{-0.5\height}{\includegraphics[height=\rowH]{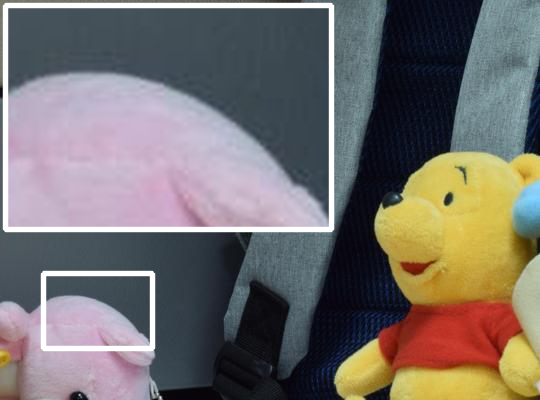}} &
  \raisebox{-0.5\height}{\includegraphics[height=\rowH]{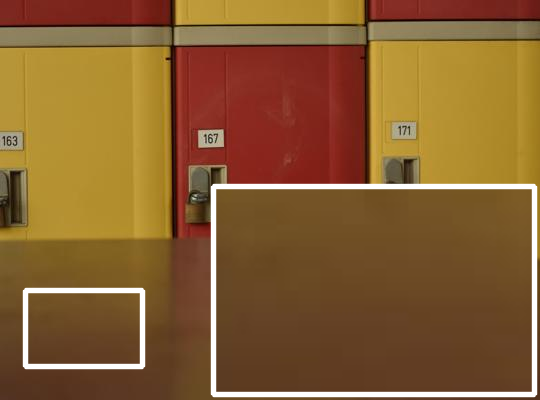}} \\

\parbox[c][\rowH][c]{0pt}{\makebox[0pt][r]{\labelsize\rotatebox{90}{WindowSeat}}} &
  \raisebox{-0.5\height}{\includegraphics[height=\rowH]{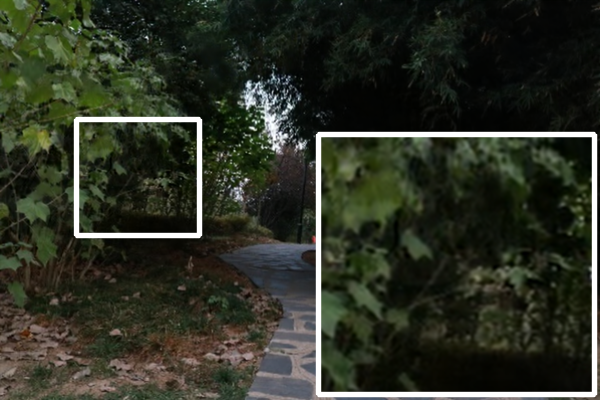}} &
  \raisebox{-0.5\height}{\includegraphics[height=\rowH]{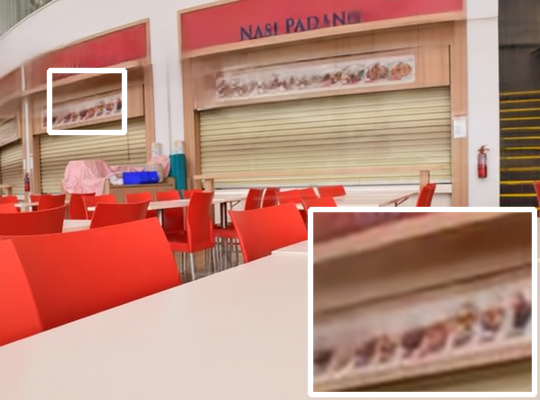}} &
  \raisebox{-0.5\height}{\includegraphics[height=\rowH]{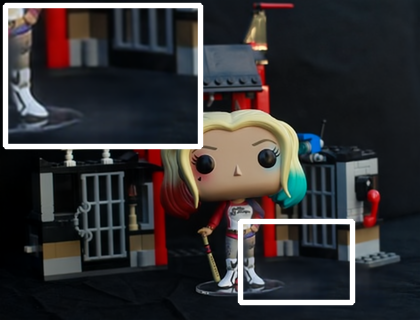}} &
  \raisebox{-0.5\height}{\includegraphics[height=\rowH]{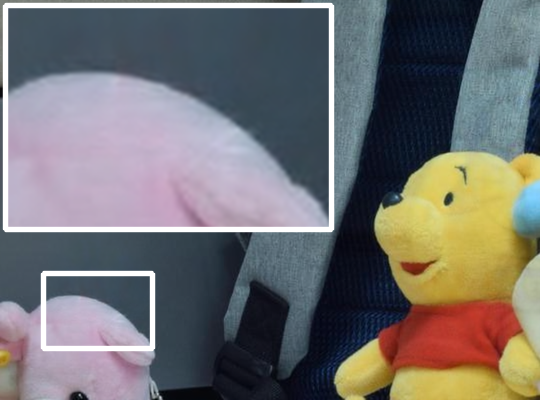}} &
  \raisebox{-0.5\height}{\includegraphics[height=\rowH]{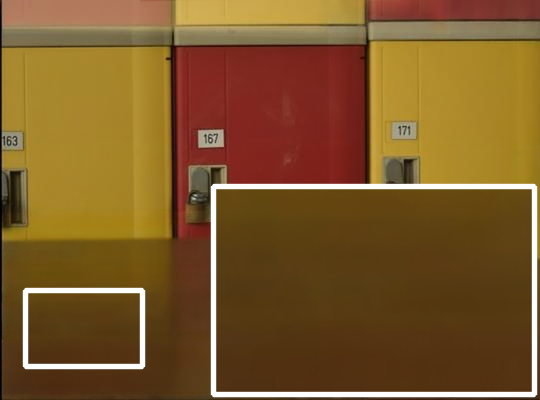}} \\

\parbox[c][\rowH][c]{0pt}{\makebox[0pt][r]{\labelsize\rotatebox{90}{DAI}}} &
  \raisebox{-0.5\height}{\includegraphics[height=\rowH]{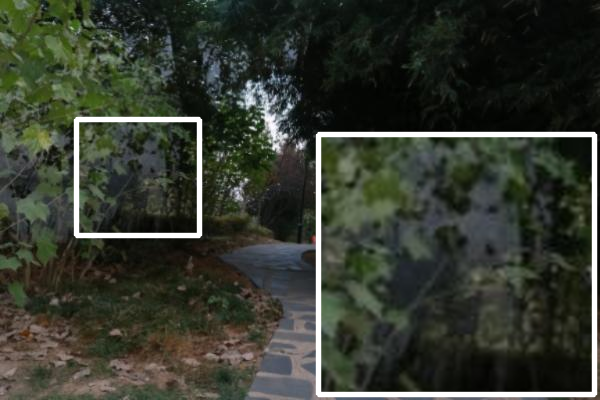}} &
  \raisebox{-0.5\height}{\includegraphics[height=\rowH]{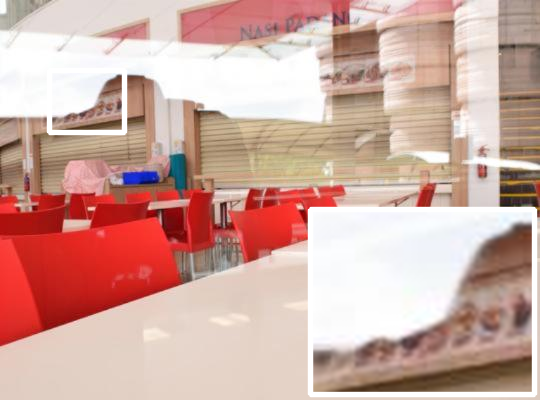}} &
  \raisebox{-0.5\height}{\includegraphics[height=\rowH]{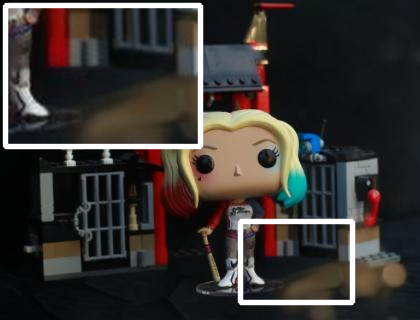}} &
  \raisebox{-0.5\height}{\includegraphics[height=\rowH]{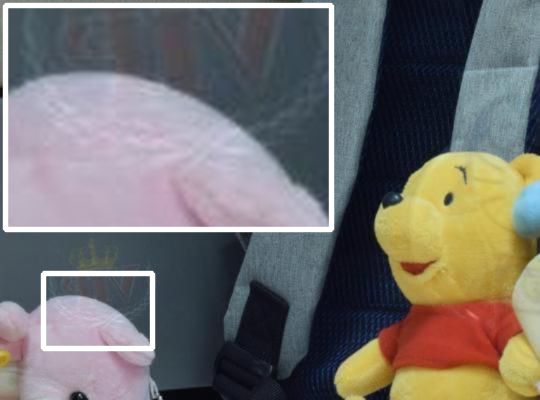}} &
  \raisebox{-0.5\height}{\includegraphics[height=\rowH]{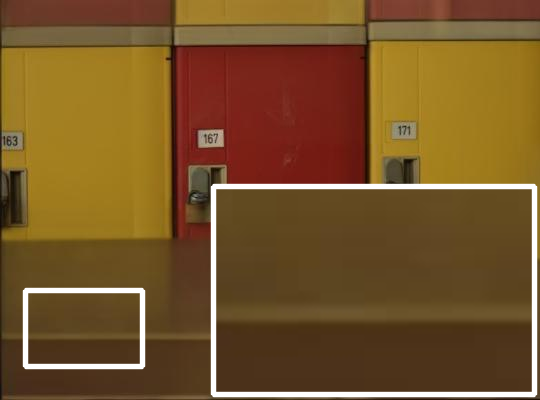}} \\
  
\parbox[c][\rowH][c]{0pt}{\makebox[0pt][r]{\labelsize\rotatebox{90}{DSIT}}} &
  \raisebox{-0.5\height}{\includegraphics[height=\rowH]{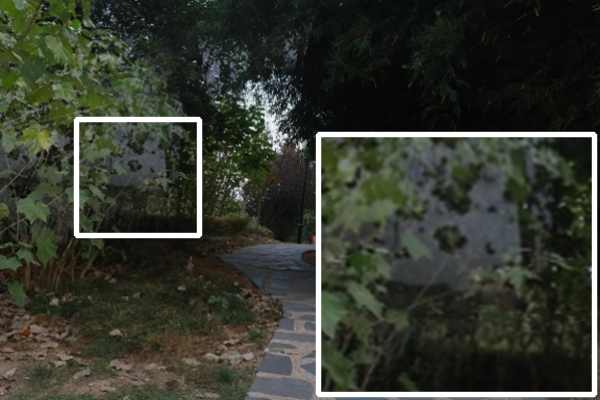}} &
  \raisebox{-0.5\height}{\includegraphics[height=\rowH]{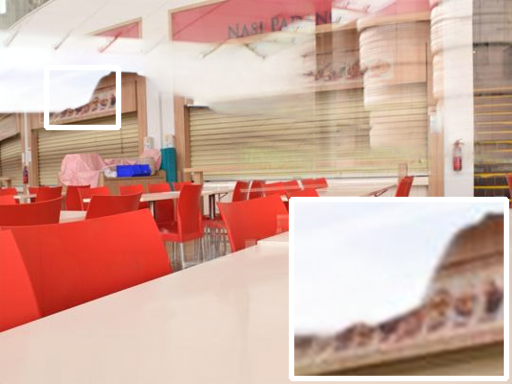}} &
  \raisebox{-0.5\height}{\includegraphics[height=\rowH]{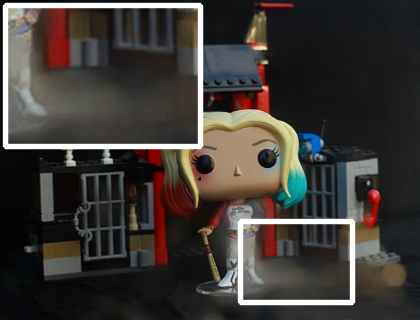}} &
  \raisebox{-0.5\height}{\includegraphics[height=\rowH]{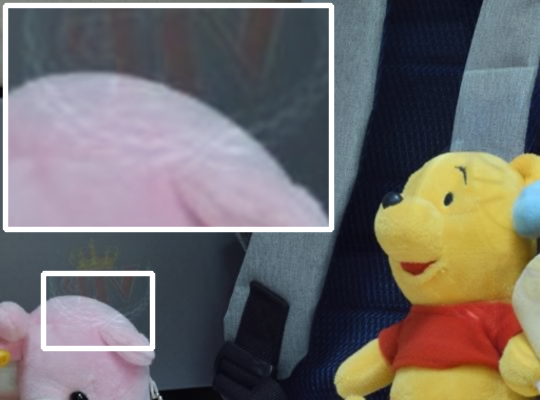}} &
  \raisebox{-0.5\height}{\includegraphics[height=\rowH]{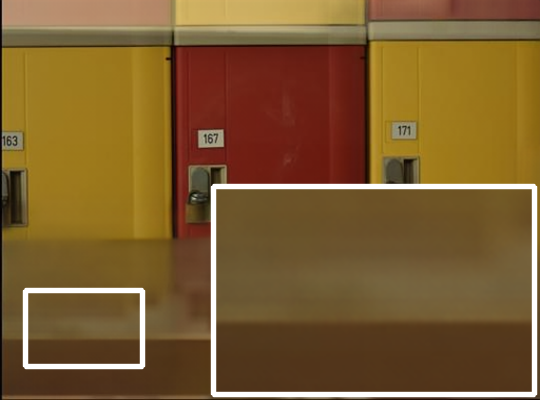}} \\
  
\parbox[c][\rowH][c]{0pt}{\makebox[0pt][r]{\labelsize\rotatebox{90}{DSRNet}}} &
  \raisebox{-0.5\height}{\includegraphics[height=\rowH]{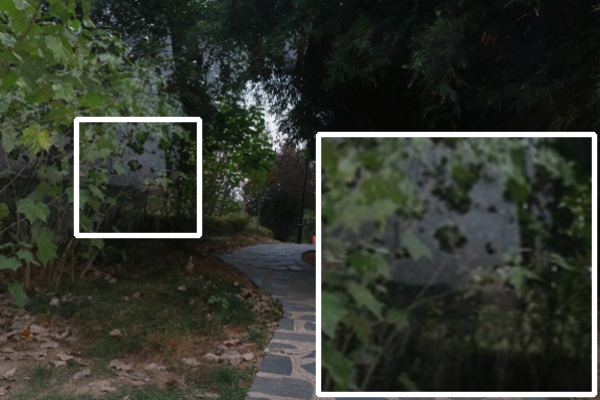}} &
  \raisebox{-0.5\height}{\includegraphics[height=\rowH]{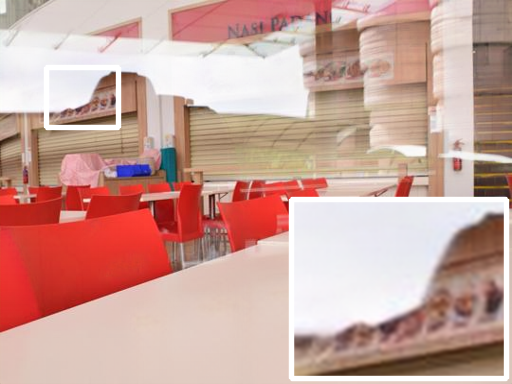}} &
  \raisebox{-0.5\height}{\includegraphics[height=\rowH]{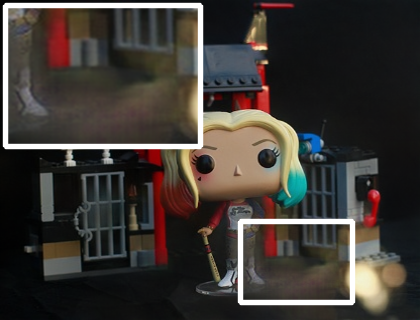}} &
  \raisebox{-0.5\height}{\includegraphics[height=\rowH]{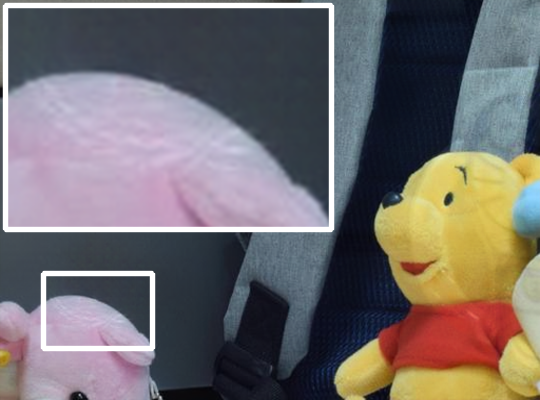}} &
  \raisebox{-0.5\height}{\includegraphics[height=\rowH]{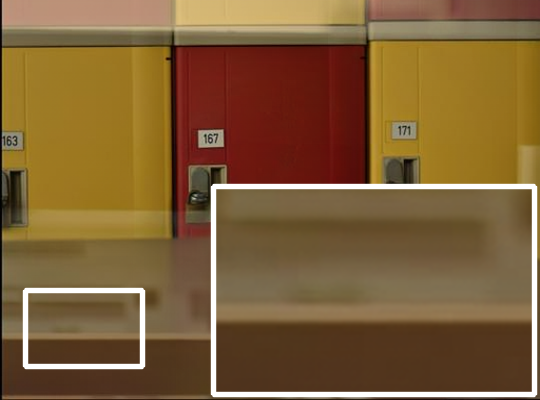}} \\

\parbox[c][\rowH][c]{0pt}{\makebox[0pt][r]{\labelsize\rotatebox{90}{RDNet}}} &
  \raisebox{-0.5\height}{\includegraphics[height=\rowH]{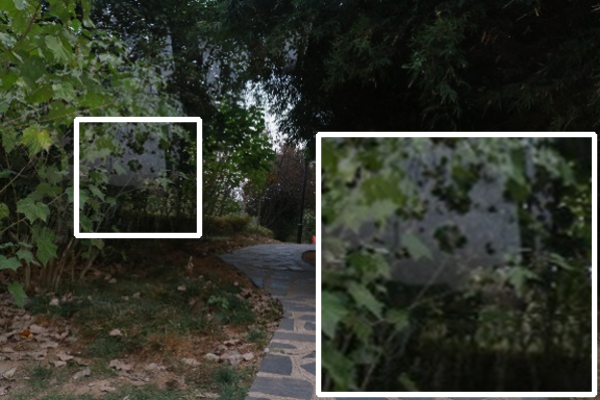}} &
  \raisebox{-0.5\height}{\includegraphics[height=\rowH]{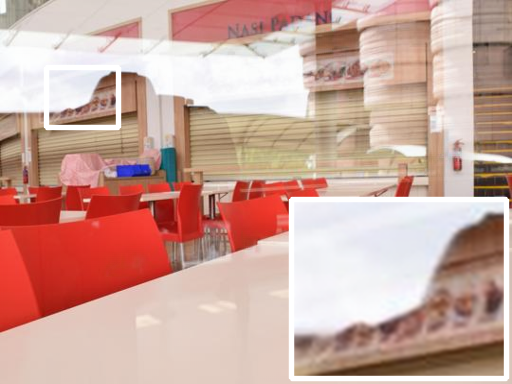}} &
  \raisebox{-0.5\height}{\includegraphics[height=\rowH]{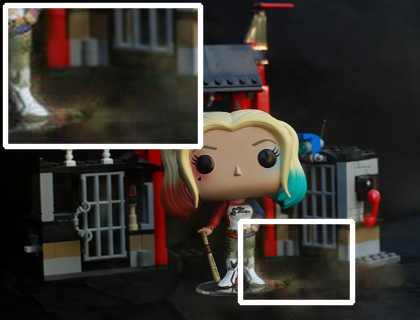}} &
  \raisebox{-0.5\height}{\includegraphics[height=\rowH]{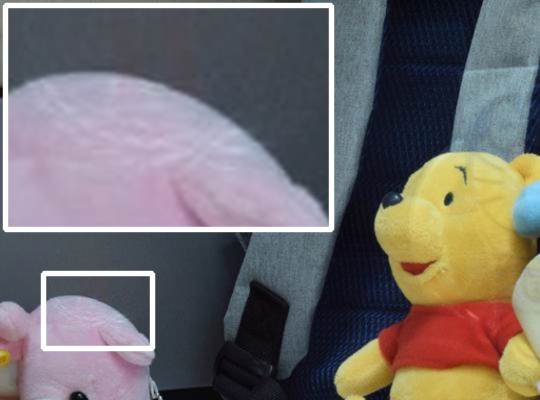}} &
  \raisebox{-0.5\height}{\includegraphics[height=\rowH]{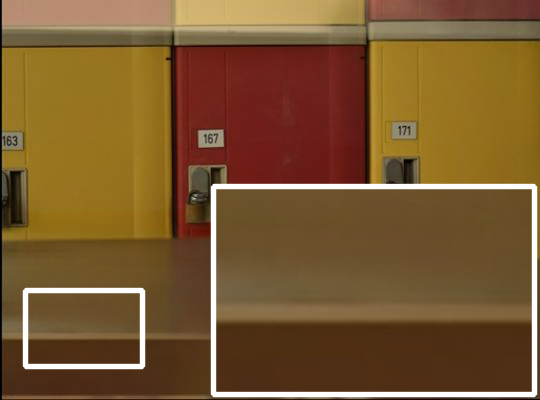}} \\

\end{tabular}

\caption{\textbf{Qualitative comparison.} Each column shows one Input-GT pair with the corresponding predictions from \ourmodel and SotA methods. \ourmodel detects and removes the reflection in the first two examples, while other methods leave the reflections unaltered. Columns 3-5 visualize the improved reflection removal capabilities of \ourmodel, leaving fewer artifacts in the predictions. Best viewed on screen and zoomed in. }
\label{fig:qual_comparison}

\end{figure*}

\begin{table*}[t]
\centering
\caption{ \textbf{Ablation on training data} evaluated on \textit{Nature} (20), \textit{Real} (20), and \textit{SIR}$^2$ sub-datasets (\textit{Postcard} (199), \textit{Objects} (200), \textit{Wild} (55)). }
\label{tab:ablation_alpha_vs_pbr}
\setlength{\tabcolsep}{4pt}
\renewcommand{\arraystretch}{1.12}
\begin{adjustbox}{max width=0.85\textwidth}
\begin{tabular}{cc *{10}{c}}
\toprule
\multirow{2}{*}{Alpha-Blended} &
\multirow{2}{*}{PBR} &
\multicolumn{2}{c}{\textit{Nature} (20)} &
\multicolumn{2}{c}{\textit{Real} (20)} &
\multicolumn{2}{c}{\textit{Objects} (200)} &
\multicolumn{2}{c}{\textit{Postcard} (199)} &
\multicolumn{2}{c}{\textit{Wild} (55)} 
\\ 
\cmidrule(lr){3-4}\cmidrule(lr){5-6}\cmidrule(lr){7-8}\cmidrule(lr){9-10}\cmidrule(lr){11-12}
\multicolumn{1}{c}{} & \multicolumn{1}{c}{} & 
PSNR $\uparrow$ & SSIM $\uparrow$
& PSNR $\uparrow$ & SSIM $\uparrow$
& PSNR $\uparrow$ & SSIM $\uparrow$
& PSNR $\uparrow$ & SSIM $\uparrow$
& PSNR $\uparrow$ & SSIM $\uparrow$ \\
\midrule 
\checkmark & \checkmark 
    & 26.96 & 0.845 & \underline{26.45} &\textbf{ 0.860} & \underline{28.62} & \underline{0.939} & \underline{27.81} & \underline{0.930} & 28.89 & \underline{0.932} \\
\checkmark & \xmark
& \textbf{27.25} & \textbf{0.850} & \textbf{26.53 } & \underline{0.859} & 27.80 & 0.931 & 27.39 & \underline{0.930} & \underline{28.93} & 0.928 \\ 
 \midrule
\xmark & \checkmark 
& \underline{27.12} & \underline{0.849} & 26.28 & 0.856 & \textbf{28.81} & \textbf{0.944} & \textbf{29.17} & \textbf{0.934} & \textbf{28.97} & \textbf{0.935} \\
\bottomrule
\end{tabular}

\end{adjustbox}
\end{table*}

\begin{figure}[htbp]
\centering
\resizebox{\linewidth}{!}{
\includegraphics[width=\textwidth]{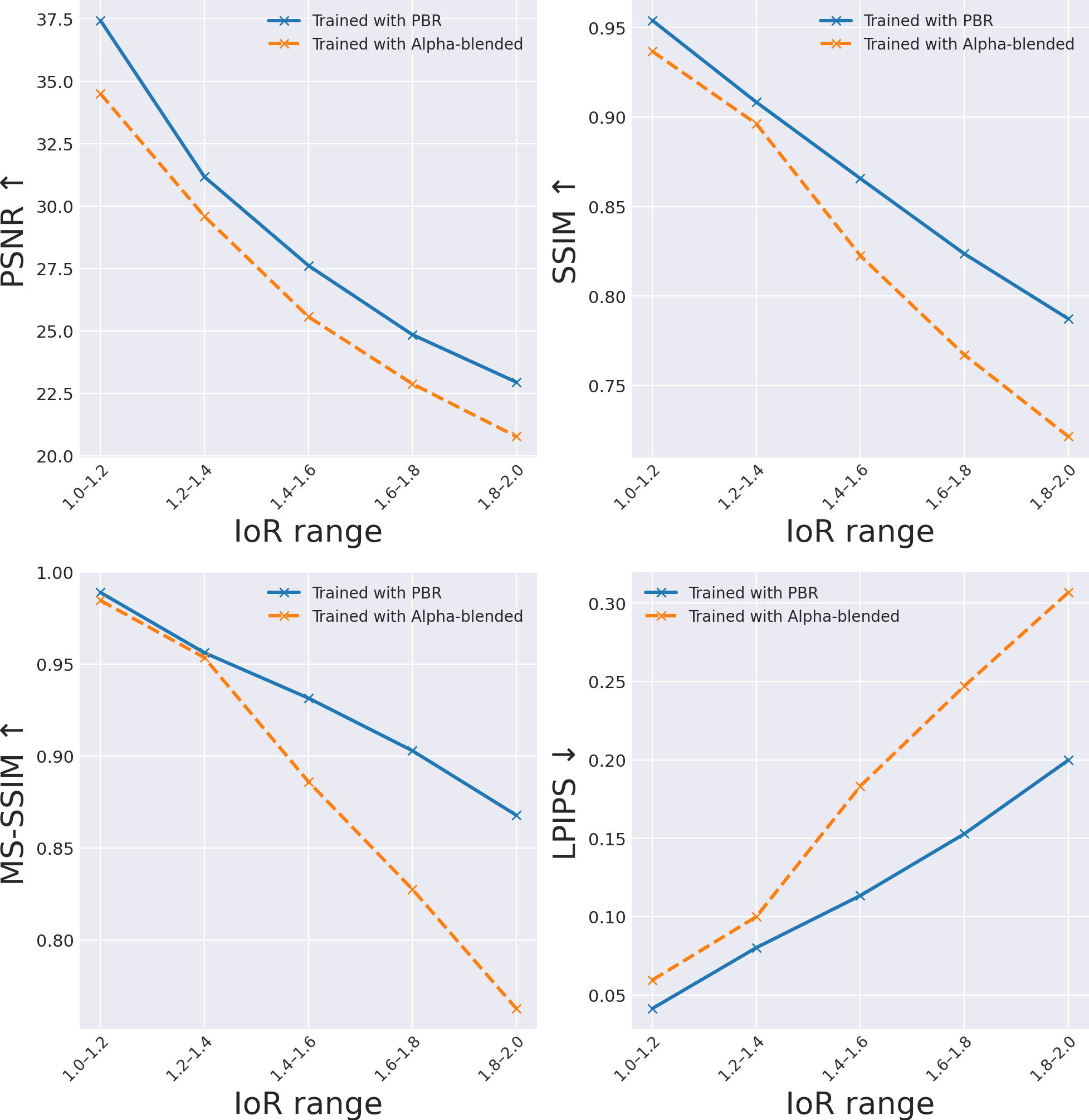}
}
\caption{\textbf{Ablation of PBR training data.} Comparing PSNR, SSIM, MS-SSIM, and LPIPS metrics of two models on five test splits. One model was trained on PBR data and the other on alpha-blended data. The x-axis denotes the IoR range of the data split, indicating increasing reflection strength. }
\label{fig:ior_bins}
\end{figure}

\subsection{Experimental Setup}

\noindent \textbf{Datasets. } Following previous work ~\cite{zhu2024revisiting, hong2024differ, hu2024single, zhao2025reversible, cai2025f2t2, huang2025single, hu2025dereflection}, we include the train split of \textit{Nature}~\cite{li2020single} and \textit{Real}~\cite{zhang2018single} in our training data and evaluate on the corresponding in-domain test splits. The \textit{SIR}$^2$~\cite{SIR2-iccv17} benchmark is evaluated in a zero-shot manner and contains three sub-datasets: \textit{Objects} (200), \textit{Postcard} (199), and \textit{Wild} (54). \textit{SIR}$^2$ (500) contains 56 additional wild-scene samples. 

\noindent \textbf{Evaluation details. } 
We compare the performance of \ourmodel against the officially reported PSNR and SSIM metrics from other SotA methods in Tab.~\ref{tab:main_benchmark}. Additionally, we re-evaluate available SotA methods \cite{hu2023single, hu2025dereflection, zhao2025reversible, hu2024single} with the best checkpoint available and compare MS-SSIM and LPIPS values in Tab.~\ref{tab:offline_evaluation}. For re-evaluation, all predictions are saved to disk and evaluated in 8-bit precision. 
SSIM and PSNR are computed with skimage, MS-SSIM with pytorch-msssim, and LPIPS with the lpips library.

\noindent \textbf{PBR data.}
We resort to the procedure described in Sec.~\ref{sec:data_synthesis} and Fig.~\ref{fig:pbr} to produce high-quality data.
Specifically, we sourced 924 HDR panoramic images from the public domain~\cite{polyhaven}, featuring diverse urban and nature settings.
The pool of sRGB images is formed from the COCO dataset~\cite{everingham2010pascal}.
For each PBR sample, the foreground is taken from the sRGB pool, and the background is sampled uniformly from both pools.
The dataset contains 25,000 rendered images in total, exhibiting varying range in reflection strengths and form. More specifically, we sample IoR values between 1.25 and 1.75, roughness between 0 and 0.05, and glass thickness between 0 and 5cm. 

\noindent \textbf{Implementation details. }
The LoRA adapter is trained for 11k steps on a single consumer GPU. 
AdamW~\cite{Loshchilov2017DecoupledWD} is initialized with a learning rate of $10^{-5}$, following 100 warm-up steps to $10^{-4}$ and a linear decay of $5 \times 10^{-6}$ every thousand steps. 
We use batch size 2 and enable gradient checkpointing for the VAE and the DiT. 
We set $\lambda_{\text{PSNR}}=0.1$ and $\lambda_{\text{SSIM}}=20$ and apply global gradient-norm clipping before each optimizer step. 
We use a LoRA adapter of rank 128 with Gaussian initialization. During training, we apply random cropping and color jitter, which includes brightness, contrast, saturation, and hue augmentations.

\vspace{1ex}
\noindent \textbf{Model quantization.}
A detailed overview of the number of trainable, frozen, and quantized parameters is provided in Tab.~\ref{tab:param_counts_compact}. To fine-tune a 12.5B-parameter model on a consumer GPU within one day, we follow the quantization scheme of ~\cite{liu2025fluxqlora} and adapt it to our DiT-based architecture~\cite{batifol2025flux}. The VAE, as well as the first and last layers of the DiT, are kept in \texttt{bfloat16}, while 95.7\% of the total parameters are quantized to 4-bit. We also quantize the first and second moments of AdamW~\cite{Loshchilov2017DecoupledWD} using the \texttt{bitsandbytes}~\cite{dettmers2022optimizers} library and use \texttt{PEFT}~\cite{peft} to attach \texttt{bfloat16} LoRA~\cite{hu2022lora} adapters, which we then train with the AdamW optimizer in a memory-efficient fine-tuning setup. The loss computations are performed in \texttt{float32}. Training with batch size 2 and training resolution of 608 has a peak GPU memory consumption of 21 GB.

\vspace{1ex}
\noindent \textbf{Resolution strategy. }
To preserve aspect ratio and match the training setup, we split the image into overlapping tiles and resize each tile to the training resolution ($608 \times 608$) using Lanczos resampling. If the input resolution is lower than the training resolution, we first upsample the shorter side of the input image and then apply tiling. In regions where tiles overlap, the final prediction is computed as a linear combination. 
Multiple tiles of an image are stacked and then processed in parallel. 

\noindent \textbf{Apache 2.0 model.}
The open-source model we provide (ours, Apache 2.0) in Tab.~\ref{tab:main_benchmark},~\ref{tab:offline_evaluation} is based on Qwen Image-Edit-2509~\cite{wu2025qwenimagetechnicalreport}.
It was trained with a batch size 1 and a processing resolution of 768$\times$768.

\subsection{Comparison with the State of the Art}
We compare our method against prior state-of-the-art approaches in Tab.~\ref{tab:main_benchmark} and report PSNR and SSIM values as provided in the respective papers; missing values are marked with ``---''.
Our method achieves a 1.56 dB gain in PSNR and raises the SSIM score from 0.930 to 0.940 on the zero-shot SIR2 (500) dataset. 
On the in-domain Real dataset~\cite{zhang2018single} we achieve a PSNR gain of 1.06 dB and an SSIM improvement from 0.846 to 0.856. On Nature~\cite{li2020single}, our method is on par with~\cite{huang2025single}, which can be explained by slight pixel-shifts present between input and GT. \newline
Additionally, we compare MS-SSIM and LPIPS with the best open-source checkpoints from DAI \cite{hu2025dereflection}, RDNet \cite{zhao2025reversible}, DSIT \cite{hu2024single}, and DSRNet \cite{hu2023single} in Tab.~\ref{tab:offline_evaluation}. Quantitative MS-SSIM and LPIPS results further confirm that \ourmodel achieves perceptually better reflection removal. \newline
We qualitatively evaluate \ourmodel in Fig.~\ref{fig:qual_comparison}. Our method identifies reflections more reliably and removes both simple and complex reflections with fewer artifacts than competing approaches. In regions with very strong reflections, \ourmodel further exhibits strong inpainting capabilities, inherited from the base DiT.
\begin{figure}[t!]
\centering
\resizebox{\linewidth}{!}{
\includegraphics[width=\linewidth]{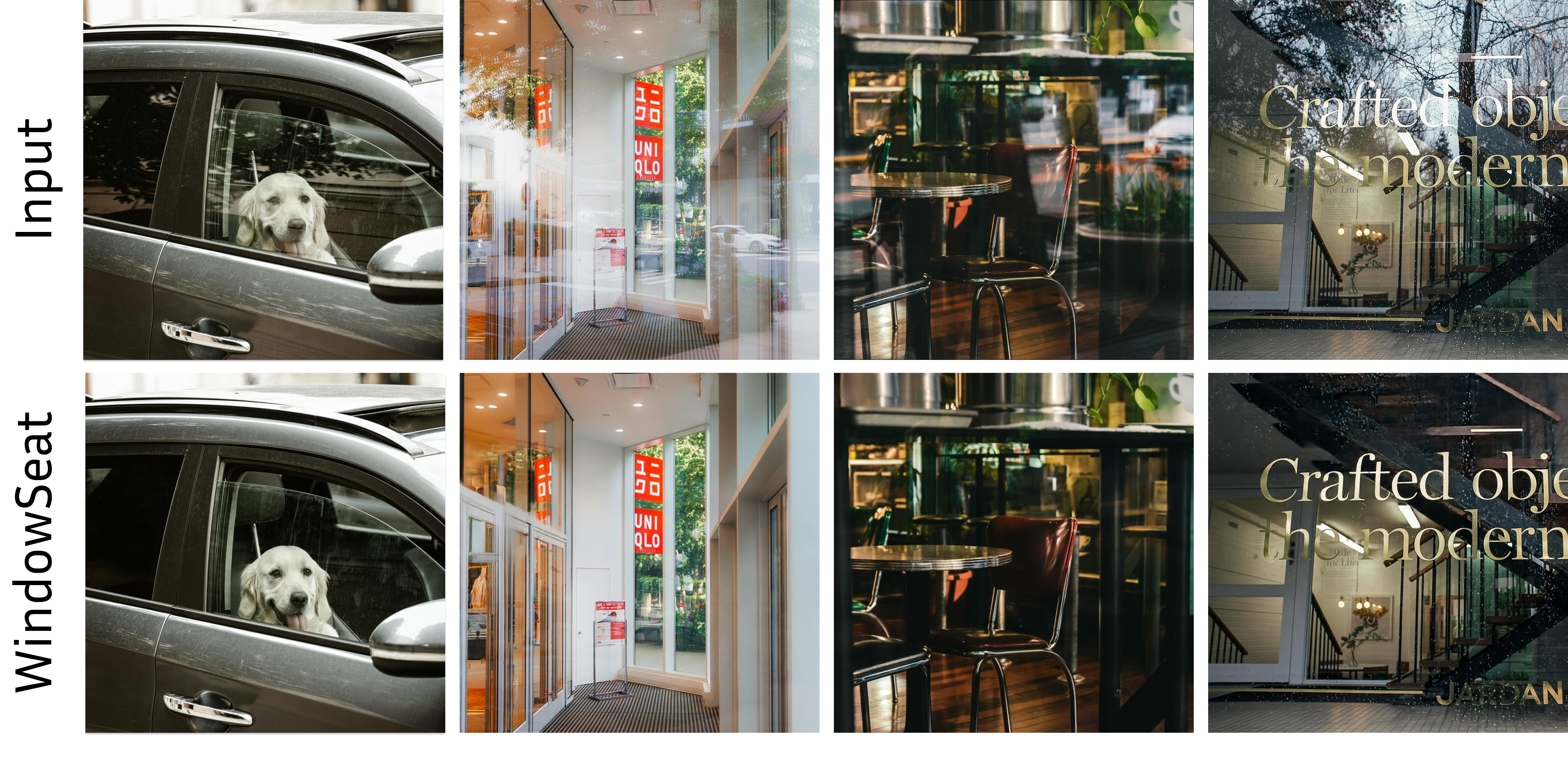}
}
\caption{
\textbf{\ourmodel predictions in the wild.} Our model demonstrates generalization across real-world reflection-removal scenarios. It effectively handles previously unseen conditions and remains robust even under challenging circumstances, including intense reflections and extreme cases such as wet or contaminated glass surfaces.
}
\label{fig:in_the_wild_pred}
\end{figure}

\begin{figure}[t!]
\centering
\resizebox{\linewidth}{!}{
\includegraphics[width=\linewidth]{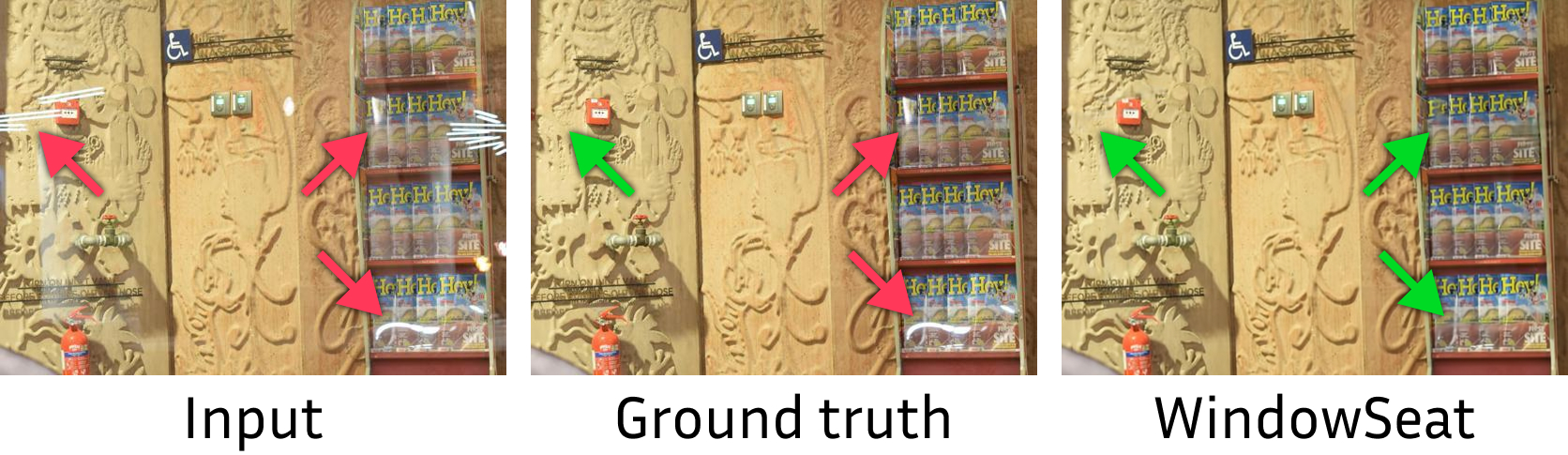}
}
\caption{
\textbf{\ourmodel failures.}
Ground truth of this SIR$^2$ (WildScene 010) sample contains secondary reflections, removed by our method.
Overshooting with removal may lead to loss of realism.
}
\label{fig:worse_gt}
\end{figure}

\subsection{Ablation Study}
\label{sec:ablation}
We ablate the network architecture, loss functions, and training data to quantify the contribution of each design choice.

\vspace{1ex}
\noindent \textbf{Architecture. }
The base DiT image editing model stacks two latent inputs along the token dimension, namely the encoded image latent $\mathbf{z}_{\!B}=\mathcal{E}(\mathbf{B})$ and noise. Since the GPU requirements grow quadratically with token length, we experiment whether and which second token set is best for one-step reflection removal. 
In Tab.~\ref{tab:ablation_pred_z1_z2}, we show the results of re-training three models with different latent inputs and modes, revealing the following insights. 

Passing two latents to the network instead of only $\mathcal{E}(\mathbf{B})$ increases the peak GPU memory consumption from 19GB to 21GB. However, it significantly improves the performance compared to omitting the second token set, up to almost 3 dB in the \textit{Postcard} test set. 
Using $\mathcal{N}(0, \mathbf{I})$ as the second latent slightly increases PSNR in \textit{Nature}, but overall performs worse than simply duplicating the RGB latent $\mathcal{E}(\mathbf{B})$. 

\newcommand{\zedit}{\hat{\mathbf{z}}_{\text{edit}}}
\newcommand{\vel}{\Delta \mathbf{z}_\theta}
\newcommand{\tedit}{t_{\text{edit}}}

Third, we ablate the output parameterization of the DiT. In the \emph{flow}
variant, the model predicts a latent-space velocity (change in latent space)
$\mathbf{v}_\theta$. In the \emph{latent} variant, the model directly predicts the
reflection-free latent $\hat{\mathbf{z}}_{\text{edit}}$.
Predicting the latent-space velocity $\mathbf{v}_\theta$ rather than the absolute latent $\hat{\mathbf{z}}_{\text{edit}}$ consistently yields better inpainting,
an example of which is shown in Fig.~\ref{fig:ablation_pbr_and_obj} (right).

\vspace{1ex}
\noindent \textbf{Loss. }
We use a PSNR-based loss to reduce global pixel-wise reconstruction error and an SSIM loss to preserve local structure and perceptual quality. 
Tab.~\ref{tab:ablation_losses} summarizes the impact of the two pixel losses $\mathcal{L}_{\mathrm{PSNR}}$ and $\mathcal{L}_{\mathrm{SSIM}}$. While omitting $\mathcal{L}_{\mathrm{SSIM}}$ still yields competitive results in \textit{Real} (20), \textit{Nature} (20), and \textit{Postcard} (199), omitting $\mathcal{L}_{\mathrm{PSNR}}$ is significantly worse in PSNR (up to 7 dB on Wild) and SSIM metrics. Overall, using both losses is best. 

\vspace{1ex}
\noindent \textbf{PBR data.}
We systematically analyze the effect of the physically based rendered data qualitatively and quantitatively. Concretely, we train once with an alpha-blended augmentation, once with our PBR dataset, and once with a mixture of both and then analyze the differences. 
To make a fair comparison, all experiments sample their transmission and blended layers from the same dataset, namely COCO \cite{lin2014microsoft}. 
Tab.~\ref{tab:ablation_alpha_vs_pbr} demonstrates that training without alpha-blended data achieves the best results for SIR$^2$ with a 1.36 dB gain in \textit{Postcard} (199). Fig.~\ref{fig:ablation_pbr_and_obj} (left) shows an example where training on alpha-blended data fails, while the model trained on PBR data successfully removes the reflection. 

To further analyze the effect of physically based rendering, we created a test set of 30 scenes, rendered in five settings with progressively increasing glass IoR values, resulting in progressively stronger reflections. Fig.~\ref{fig:ior_bins} presents the behaviour of PSNR, SSIM, MS-SSIM, and LPIPS metrics across these test sets, demonstrating that training on PBR data provides improved robustness to stronger reflections.

\subsection{In-the-wild Evaluations and Limitations}
\label{sec:in_the_wild_and_lims}
We show qualitative in-the-wild predictions in Fig.~\ref{fig:in_the_wild_pred}, demonstrating the strong generalization capacity of \ourmodel.
However, as shown in Fig.~\ref{fig:worse_gt}, it removes secondary and higher-order reflections, which might be misaligned with ground truth or the user's intent. Future directions might include more fine-grained user control, \eg specifying how many reflection layers should be removed.

\section{Conclusion}
\label{sec:conclusion}

We presented \ourmodel, a foundation DiT-based model fine-tuned with LoRA that achieves efficient single-image reflection removal on a single consumer GPU within one day of training.
Unlike prior work, we place a strong emphasis on a high-fidelity PBR data generation pipeline, enabling a model that attains state-of-the-art performance on both established benchmarks and in-the-wild images.
These results demonstrate that diffusion transformers, supported by physically grounded synthetic data and lightweight adaptation, provide an efficient and scalable framework for reflection removal and a strong basis for advancing other computational photography tasks. 
Looking ahead, the same principles can be extended to more challenging settings, including temporally consistent video and reflections involving subject self-shadows, paving the way towards more comprehensive through-glass imaging.

{
    \small
    \bibliographystyle{ieeenat_fullname}
    \bibliography{main}
}

\end{document}